\PassOptionsToPackage{table,dvipsnames}{xcolor}
\documentclass[10pt,twocolumn,letterpaper]{article}

\usepackage{cvpr}              

\definecolor{cvprblue}{rgb}{0.21,0.49,0.74}
\usepackage[pagebackref,breaklinks,colorlinks,allcolors=cvprblue]{hyperref}

\usepackage[accsupp]{axessibility}  

\usepackage{amsmath}

\newtheorem{theorem}{Theorem}
\newtheorem{assumption}[theorem]{Assumption}
\newenvironment{proof}{\par\noindent{\bf Proof\ }}

\usepackage{graphicx}
\usepackage{booktabs}
\usepackage{multirow}
\usepackage{soul}

\usepackage{math_commands}

\def\paperID{17681} 
\def\confName{CVPR}
\def\confYear{2025}

\title{Where's the liability in the Generative Era?\\Recovery-based Black-Box Detection of AI-Generated Content}

\author{Haoyue Bai\\
University of Wisconsin-Madison\\
{\tt\small haoyue.bai@wisc.edu}
\and
Yiyou Sun\\
University of California, Berkeley\\
{\tt\small sunyiyou@berkeley.edu}
\and
Wei Cheng\\
NEC Laboratories America\\
{\tt\small weicheng@nec-labs.com}
\and
Haifeng Chen\\
NEC Laboratories America\\
{\tt\small haifeng@nec-labs.com}
}

\begin{document}
\maketitle
\begin{abstract}

The recent proliferation of photorealistic images created by generative models has sparked both excitement and concern, as these images are increasingly indistinguishable from real ones to the human eye. While offering new creative and commercial possibilities, the potential for misuse, such as in misinformation and fraud, highlights the need for effective detection methods. Current detection approaches often rely on access to model weights or require extensive collections of real image datasets, limiting their scalability and practical application in real-world scenarios. In this work, we introduce a novel black-box detection framework that requires only API access, sidestepping the need for model weights or large auxiliary datasets. Our approach leverages a corrupt-and-recover strategy: by masking part of an image and assessing the model’s ability to reconstruct it, we measure the likelihood that the image was generated by the model itself. For black-box models that do not support masked-image inputs, we incorporate a cost-efficient surrogate model trained to align with the target model’s distribution, enhancing detection capability. Our framework demonstrates strong performance, outperforming baseline methods by 4.31\% in mean average precision across eight diffusion model variant datasets. Code is publicly available at \url{https://github.com/HaoyueBaiZJU/genai-detect}.

\end{abstract}

\section{Introduction}
\label{sec:intro}

The rapid advancement of generative models~\cite{ramesh2021zero, rombach2022high, betker2023improving} has driven remarkable progress in synthesizing photorealistic images, offering numerous benefits yet also raising concerns about potential misuse. For instance, the creation of fake images, such as the widely circulated ``Trump getting arrested'' photo~\cite{bbc_news_2024}, can escalate public confusion and fuel misinformation. Similarly, a Hong Kong employee was deceived into transferring money to criminals through an AI-generated video call~\cite{guardian2024deepfake}. 

This underscores the urgent need for robust methods to distinguish real images from AI-generated ones. However, developing such methods is challenging given that current generative models can produce images with a photorealistic quality. For instance, Appendix~\ref{App:hard} 
showcases examples where humans struggle to accurately differentiate real from fake, with accuracy often as low as chance level (50-50). In a curated dataset of challenging cases, humans could only identify 70\% of the fake images correctly as we show in Section~\ref{exp:human}.

\begin{figure}
\centering
\includegraphics[width=0.48\textwidth]{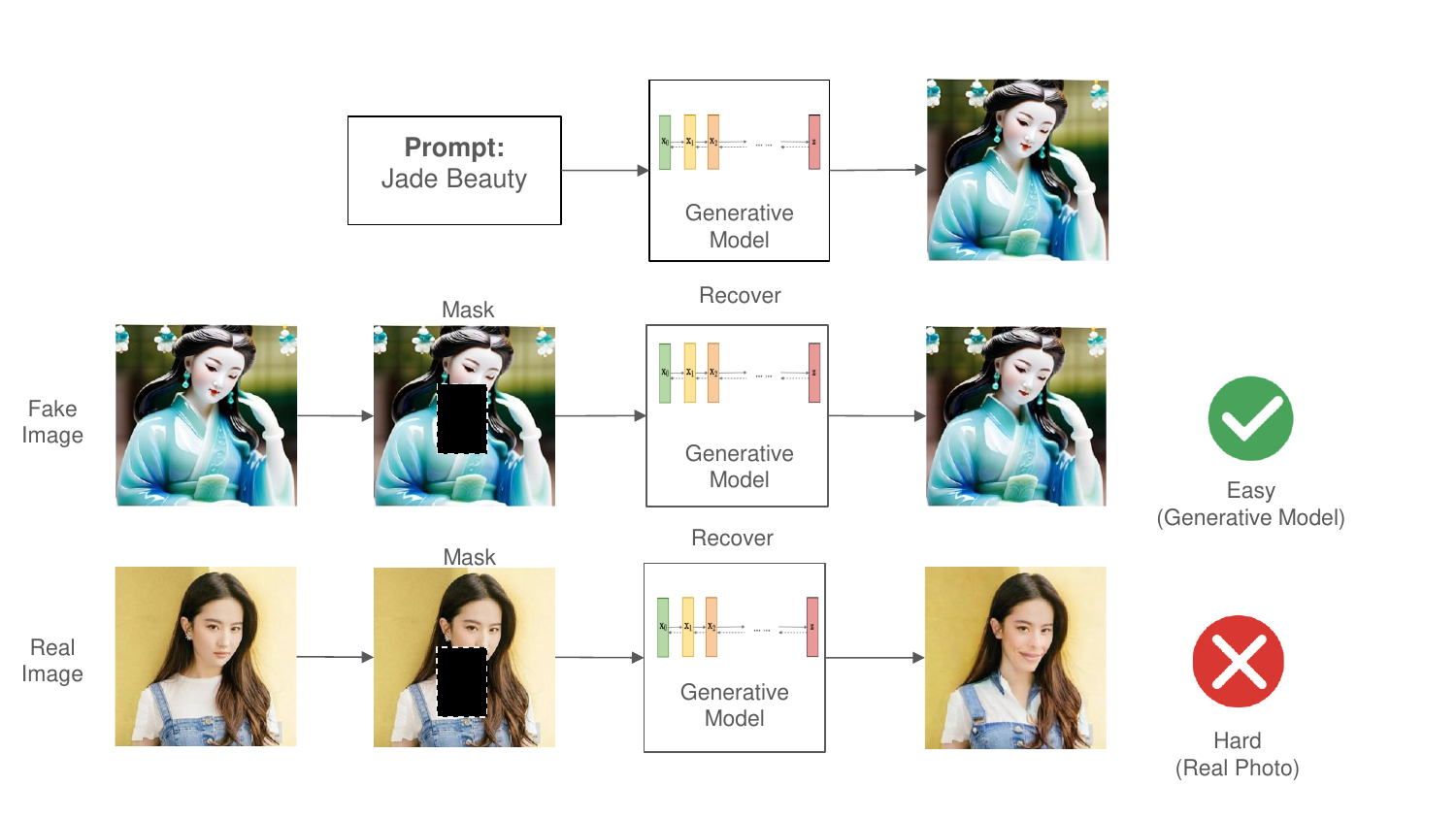}
\caption{Visual quality comparison of surrogate model recovery output between real images and images generated by the target model from masked examples. We observe that images generated by target models with masks are more likely to be accurately recovered compared to real images. These results are based on Stable Diffusion as the surrogate model.}
\end{figure}\label{fig:teaser}

Considerable effort has gone into developing detection methods~\cite{wang2024did, chai2020makes, laszkiewicz2023single, graham2023denoising}. 
However, many of these are "white-box" methods that require access to model weights or token information. The real challenge, however, lies in detecting fake images generated through widely accessible, low-cost "black-box" APIs~\cite{dalle3, team2024gemini}. 
In these model-agnostic or black-box settings, many other detection approaches~\cite{wang2020cnn, ojha2023towards} 
typically rely on binary classifiers trained on limited datasets of real versus fake images. However, given the vast diversity of real-world images, such classifiers are prone to overfitting and will exhibit overconfidence or vulnerability when exposed to previously unseen data~\cite{bai2023feed} 
in open-world environments. As diffusion-based generative models continue to evolve, there remains a critical need for practical, generalizable, interpretable, and robust methodologies to reliably detect AI-generated images.

To address these limitations, our work introduces a novel black-box fake detection method that requires only API access, without needing direct access to model weights or auxiliary datasets. Our framework is grounded in a straightforward intuition:

\begin{center} \textit{A generative model should more easily recover its\\own generated images when corrupted than it would\\with real images.} \end{center}

This intuition, validated in previous work~\cite{yangdna}
, is based on the premise that the distribution of AI-generated content inside the network significantly differs from that of real-world images. Using this insight, we developed a detection algorithm that follows a pipeline illustrated in Figure~\ref{fig:teaser}: (1) corrupt the target image by applying a mask, (2) use the generative model to recover the masked content, and (3) compare the quality of the recovery to the uncorrupted version. Ideally, if the target model generated the image, it should easily recover the masked areas, given that the image aligns with its own distribution. For example, a model that generates an image of a "clay" Trump can likely recover clay-specific features, such as a clay mouth, but would struggle to accurately reconstruct details in a realistic photo of Trump. Thus, we can discern real versus fake by calculating a measurable score that compares the original image to the corrupt-recover version. We expect this score to be higher for real images and lower for fake ones.

For public models where the API only supports generation from scratch (without a masked-image input), we employ a surrogate model. This surrogate is trained in a cost-effective manner to closely align its probability distribution with that of the target black-box model, effectively synchronizing their distributions for accurate detection.
Unlike previous methods~\cite{ojha2023towards},
which require a large corpus of real images (approximately 400k
) and significant computational resources, our approach requires fewer than 1,000 samples from the target model's API and less than 2 GPU hours of compute time.

Our main contributions are summarized as follows:

\begin{itemize}
\item \textbf{Competitive Black-Box Detection Framework.} We introduce a novel black-box detection framework that identifies AI-generated images using only API access, eliminating the need for access to model weights or large auxiliary datasets. 
\item \textbf{Novel Corrupt-and-Recover Detection Paradigm.} Our approach leverages a unique corrupt-and-recover pipeline, where a model’s ability to reconstruct its own generated content provides an effective measure to differentiate between real and fake images.
\item \textbf{Efficient Resource Requirements.} Unlike prior methods that require extensive datasets and computational resources, our approach achieves high detection performance with fewer than 1,000 API samples and minimal GPU time.
\item \textbf{Improved Generalizability and Practicality}. Our method demonstrates enhanced generalizability across diverse generative models, making it a practical and robust solution as generative models continue to improve in quality and accessibility.
\end{itemize}

\section{Related Works}

\subsection{Synthetic Image Generation}

The field of generative models has been significantly advanced with the introduction of Generative Adversarial Networks (GANs) \cite{choi2018stargan, zhu2017unpaired, brock2018large}. Some works have exploited Transformers to enhance generated image quality \cite{parmar2018image, zhang2022styleswin, chang2023muse}. The advent of diffusion models has led to significantly improved cutting-edge generation models, including Stable Diffusion \cite{rombach2022high}, DALL-E \cite{ramesh2022hierarchical}, DALL-E 2 \cite{ramesh2022hierarchical}, DALL-E 3 \cite{betker2023improving}, GLIDE \cite{nichol2021glide}, and others \cite{dhariwal2021diffusion, zhu2023conditional}.

\subsection{Detection of AI-Generated Content}

With the proliferation of synthetic image generators, designing methods to detect generated content has attracted much attention \cite{zhu2024genimage}. Some earlier works focus on detecting fake faces. 
The work in \cite{matern2019exploiting} leverages visual features such as eyes, teeth, and facial contours. The work of \cite{mccloskey2018detecting} examines color information related to the synthesis of RGB color channels.
Most recent detectors rely on traces that are invisible to the human eye, inherent to the generation process, and based on semantic, physical, or statistical inconsistencies. One direction identifies feature frequency artifacts for GAN-generated images \cite{marra2019gans, zhang2019detecting}. The work in \cite{yu2019attributing} studies learning GAN fingerprints for image attribution. Patchfor \cite{chai2020makes} uses classifiers with limited receptive fields to focus on local artifacts instead of global semantics of the images.

The work in \cite{corvi2023detection} shows that GAN detectors perform poorly on diffusion model-generated images. Recent detection techniques have begun studying diffusion model-based images. Synthbuster \cite{bammey2023synthbuster} investigates the inherent frequency artifacts during the diffusion process and leverages spectral analysis to highlight the artifacts in the Fourier transform of a residual image for fake detection.
Other works exploit lighting \cite{farid2022lighting} and perspective \cite{farid2022perspective} inconsistencies of DALL-E 2 generated images. DE-FAKE \cite{sha2023fake} focuses on advanced text-to-image generation models including DALL-E 2 and Stable Diffusion, and observes that incorporating prompts or generated captions into the detector improves classification. DIRE \cite{wang2023dire} observes that diffusion-generated images can be approximately reconstructed by a diffusion model while real images cannot.
The work in \cite{ricker2022towards} observes that diffusion models produce fewer detectable artifacts and are more difficult to detect compared to GANs, and explores retraining GAN detectors on diffusion model-generated images to show improved detection. The work in \cite{wang2024did} defines a reverse-engineering task for generative models and analyzes the disparities in reconstruction loss between the generated samples of the specific model and others.
The work in \cite{graham2023denoising} exploits denoising diffusion probabilistic models as denoising autoencoders and uses the resulting multi-dimensional reconstruction error to classify out-of-distribution inputs. Universal fake detector \cite{ojha2023towards} proposes using a pre-trained vision transformer with a final classification layer for fake detection of both GAN and diffusion model-generated images.
However, if a network is trained on a specific model, its performance degrades when used to detect images generated by another architecture \cite{cozzolino2018forensictransfer}. This suggests that each generation architecture contains its own peculiar traces.

\subsection{Black-box Detection}

The specific model used by the attacker is often unavailable. One approach is to train a classifier with fewer or even no fake images from the pre-trained generative model.
AutoGAN \cite{zhang2019detecting} investigates the artifacts induced by the up-sampler of GAN pipelines in the frequency domain to develop robust spectrum-based fake image classifiers.
Some recent works \cite{wang2020cnn, gragnaniello2021gan, mandelli2022detecting} claim to perform well on images from unseen generative models. However, it remains unclear whether this performance holds for images generated by diffusion models.

\section{Methodology}\label{sec:method}

\textbf{Motivation.} The core idea behind our approach is: \ul{a generative model should more easily recover its own generated images when corrupted than it would with real images}. Technically, we hypothesize that a generative model $\mathbb{G}_t$ can inherently “recognize” its own outputs, allowing it to reconstruct masked regions of its own generated images more effectively than it can for other images, such as those from the real world. This occurs because the model’s learned distribution aligns closely with the statistical properties of its own outputs, making these images easier to “fill in” when corrupted. In contrast, real images or images generated by other models follow distributions that $\mathbb{G}_t$ hasn’t explicitly learned, so it struggles to accurately restore missing content in these cases.

\textbf{Problem Setup.} 
Our task is to detect whether a given image $\mathbf{x}$ is produced by a target generative model $\mathbb{G}_t$ (where we only have API access), which can be framed as a binary classification problem where we want to determine if \( \mathbf{x} \) is generated by \( \mathbb{G}_t \) (i.e., \( y = 1 \)) or not (i.e., \( y = 0 \)):

\begin{equation}
    \hat{y} = 
    \begin{cases} 
        1, & \text{if } \delta(\mathbf{x}) < \tau \\
        0, & \text{otherwise}
    \end{cases}
\end{equation}

Here, \( \tau \) is a predefined threshold. If the discrepancy score \( \delta(\mathbf{x}) \) is below this threshold, we classify the image as being generated by \( \mathbb{G}_t \) (\( \hat{y} = 1 \)); otherwise, we classify it as not generated by \( \mathbb{G}_t \) (\( \hat{y} = 0 \)).

There are typically two different task settings for AI-generated content detection: black-box detection (with access only to input and output) and white-box detection (with additional information about the model internals). We focus on the black-box setting in this work, as large tech companies and AI research organizations often keep their most advanced models closed-source, such as DALL-E 3 and SORA by OpenAI. In these cases, we can only access the model through their API, while the underlying code and model weights remain unavailable to the public. Therefore, our goal is to improve black-box detection without any access to the target model weights.

\textbf{Preliminary on Diffusion Models.}
Diffusion Models are a group of probabilistic generative models. Since the milestone work DDPM~\cite{ho2020denoising}, there are numerious improvements with higher fidelity and diversity~\cite{rombach2022high, ramesh2022hierarchical, nichol2021glide}. A diffusion probabilistic model is a parameterized Markov chain trained using variational inference to produce samples matching the data after finite time, which gradually diffuse a sample from this distribution and then learn to reverse this diffusion process.

In the diffusion (or forward) process for DDPMs, a sample $\rvx_0$ (e.g., an image) is repeatedly corrupted by Gaussian noise in sequential steps $t=1,\dots, T$ in dependence of a monotonically increasing noise schedule $\{\beta_t\}_{t=1}^T$:
\begin{equation}
    q(\rvx_t \vert \rvx_{t-1}) = \mathcal{N}(\sqrt{1-\beta_t} \rvx_{t-1}, \beta_t \mathbf{I}) \enspace .
\end{equation}
With $\alpha_t = 1-\beta_t$ and ${\bar \alpha_t = \prod_{s=1}^t  \alpha_s}$, we can directly sample from the forward process at arbitrary times:
\begin{equation}
    q(\rvx_t \vert \rvx_0) = \mathcal{N}(\sqrt{\bar \alpha_t} \rvx_0, (1-\bar\alpha_t) \mathbf{I})\enspace.
\end{equation}
The noise schedule is typically designed to satisfy $q(\rvx_T \vert \rvx_0) \approx \mathcal{N}(\mathbf{0}, \mathbf{I})$.
During the denoising 
process, we aim to iteratively sample from $q(\rvx_{t-1} \vert \rvx_t)$ to ultimately obtain a clean image from $\rvx_T \sim \mathcal{N}(\mathbf{0}, \mathbf{I})$.
However, since $q(\rvx_{t-1} \vert \rvx_t)$ is intractable as it depends on the entire underlying data distribution, it is approximated by a deep neural network.
More formally, $q(\rvx_{t-1} \vert \rvx_t)$ is approximated by
\begin{equation} 
    p_\theta(\rvx_{t-1} \vert \rvx_{t}) = \mathcal{N}(\mu_\theta(\rvx_t,t), \Sigma_\theta(\rvx_t,t)) \enspace,
\end{equation}
where mean $\mu_\theta$ and covariance $\Sigma_\theta$ are given by the output of model (or the latter is set to a constant as shown in~\cite{ho2020denoising}).

\subsection{Recovery-based Detection Methods}

Given an input image $\rvx$, our goal is to determine whether it is synthesized by a generative model or if it is a real image. We define a mask $m$ to divide the image into two parts: the known pixels $(1 - m) \odot \rvx$ and the unknown pixels $m \odot \rvx$. As illustrated in Figure~\ref{fig:framework}, we apply a generative model to recover the unknown pixels $m \odot \rvx$ conditioned on the known pixels $(1 - m) \odot \rvx$. The surrogate \href{https://github.com/Stability-AI/StableDiffusion}{generative model} $\mathbb{G}_s$ is usually shared with an inpainting model, which provides both the ability to generate from scratch and to generate from partially masked inputs~\cite{Rombach_2022_CVPR}. 
The difference between the input $m \odot \rvx$ and the recovered $m \odot \rvx$ helps distinguish between real and generated images.
We compute a metric $\delta$ of this discrepancy gap and use it as a scoring function to classify the source image as either real or generated. In practice, we sample the recovery results $K$ times to account for the stochastic nature of the process and to obtain a more robust evaluation.

\textbf{Details of Scoring Function $\delta$.} 
Our learning framework is orthogonal, thus compatible with various metrics $\delta$ used for measuring discrepancy as a scoring function. We evaluate four different types of scoring functions in this work: \emph{i}) Peak Signal-to-Noise Ratio (PSNR), which measures the ratio of signal to noise; \emph{ii}) Structural Similarity Index (SSIM), which quantifies structural similarity; \emph{iii}) L1 distance, which measures absolute pixel-wise differences; and \emph{iv}) L2 distance, which measures squared pixel-wise differences. More details of $\delta$ can be found in Appendix~\ref{app:scores}.

As shown in Section~\ref{exp:ablation}, PSNR achieves better performance in fake image detection compared to other metrics. This may be due to the fact that fake images often contain subtle alterations, and PSNR excels at detecting small pixel-wise differences, making it highly sensitive to fine-grained changes. In contrast, L1 and L2 distances do not normalize these differences relative to the image’s dynamic range, while PSNR normalizes the error against the maximum possible pixel value, making it more interpretable and robust to variations in intensity. Additionally, SSIM focuses more on structural similarity and perceptual quality, which may cause it to overlook subtle pixel-level deviations.

\subsection{Distribution-Aligned Black-Box Detection}

For public models where the API only supports generation from scratch, we utilize a distribution-aligned surrogate model $\mathbb{G}_s$ to recover masked images generated by a target model $\mathbb{G}_t$. We observe that images generated by target models with masks can be accurately recovered by a distribution-aligned model, as shown in Figure~\ref{fig:teaser} where a ``clay trump'''s mouth can be recovered in similar style.

\begin{figure*}
  \centering
  \includegraphics[width=0.8\textwidth]
  {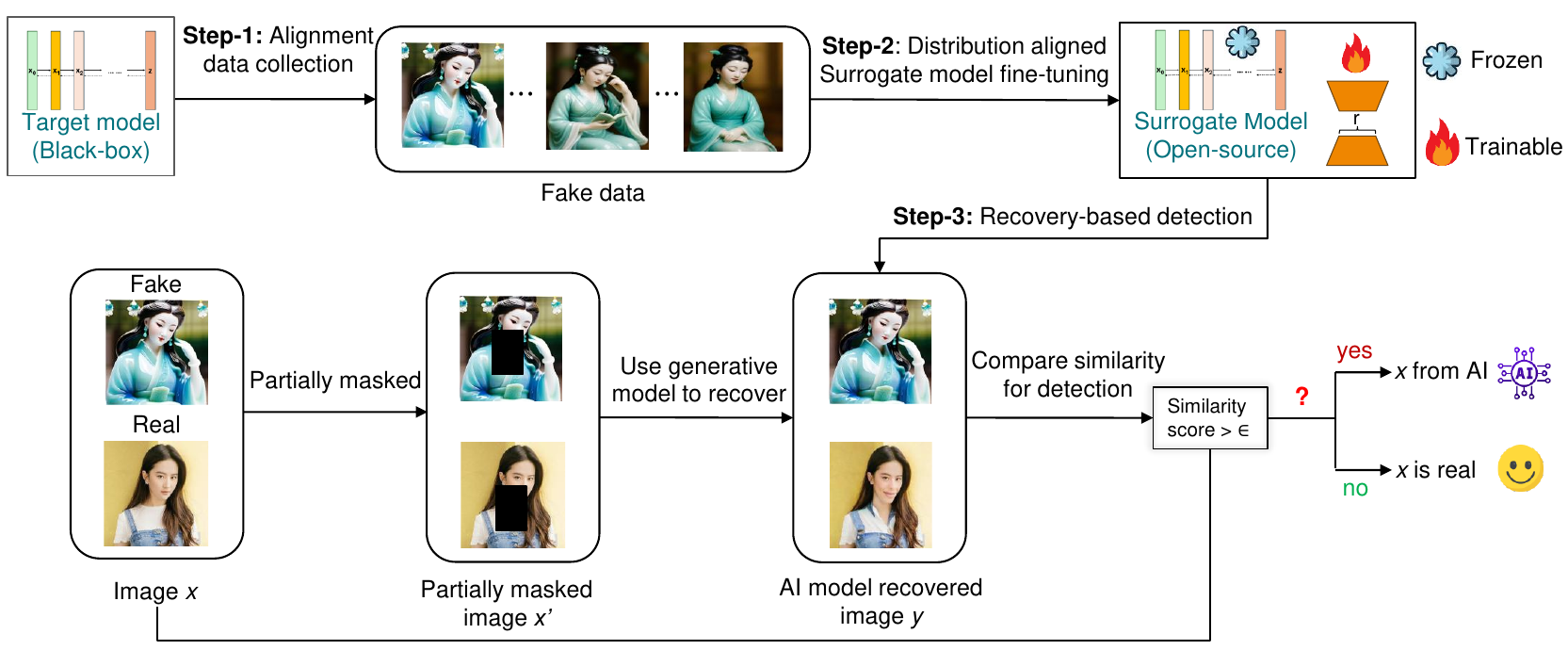}
  \caption{An overview of our proposed black-box content detection framework. Given a candidate input image, we aim to determine whether it is generated by a target model or if it is a real image. Our method first aligns the distribution of the surrogate model with that of the target model via alignment data collection and parameter-efficient surrogate model fine-tuning. We then perform recovery-based detection to calculate the scoring function for classification.}\label{fig:framework}
\end{figure*}

In the black-box detection setting, selecting an appropriate surrogate model is crucial for achieving accurate and reliable results. 
There exists a distribution gap between the given surrogate model and the target model. We aim to obtain a surrogate model that approaches the distribution of the target model by utilizing images generated by the target model. We propose a novel and efficient framework to train a distribution-aligned surrogate model that achieves good performance for black-box detection with a small-sized dataset. As shown in Figure~\ref{fig:framework}, we first collect a small set of training data generated by the target model from the publicly shared API. Then, we perform parameter-efficient fine-tuning of the surrogate model using this training dataset to align its distribution with the source model.

\textbf{Alignment Data Collection}
To align the distribution of the surrogate model $\mathbb{G}_s$ with the target model $\mathbb{G}_t$, we collect a small-sized dataset $S = \{x_i\}_{i=1}^N$ for a specific target model, referred to as the alignment dataset. Here, $N$ denotes the number of collected images, and $x_i$ refers to an image generated by the target model through publicly shared APIs. We then utilize the collected dataset $S$ to fine-tune the surrogate model $\mathbb{G}_s$, aligning its distribution with that of the target model $\mathbb{G}_t$.

\textbf{Distribution-Aligned Surrogate Model Fine-Tuning}
As shown in Figure~\ref{fig:framework}, we implement low-rank adaptation (LoRA)~\cite{hu2021lora} for the surrogate model $\mathbb{G}s$ to enable parameter-efficient fine-tuning. The LoRA model $\mathbb{G}{s+\theta}$ is trained with the collected dataset $S$, while the parameters of the original surrogate model $\mathbb{G}_s$ remain frozen.
After training, the previously misaligned model generates a distribution similar to that of the target model $\mathbb{G}_t$. Consequently, this distribution-aligned surrogate model can be utilized to perform recovery evaluation for downstream fake detection.

\subsection{Theoretical Insights}

Formally, given an input image $\rvx$, we define a mask $m$ for dividing the image into two parts: $X = (1 - m) \odot x$, and $Y_0 = m \odot x$. Next, we ask the generative model to continue generating the remaining pixels purely based on $X$, and the generated results are denoted by $Y' \sim G(\cdot|X)$. In practice, we sample the new results for $K$ times (refer to a principled choice of  $K = \Omega \big( {\sigma \log(1/\delta)}/{\Delta^2} \big)$ in Appendix~\ref{app:theory_k} ) to get a set of sequences $\Omega = \{Y_1, ..., Y_k, ..., Y_K \}$.
Our method is based on the hypothesis that the generation process $G$ of the machine typically maximizes the log probability function 
throughout the generation process, while real image creation process is different. In other words, the thought process of real images does not simply follow the likelihood maximization criterion. We find that this discrepancy between machine and real is especially enormous when conditioned on the input pixels $X$, and we state this hypothesis formally as:
\paragraph{Likelihood-Gap Hypothesis.} The expected log-likelihood of the machine generation process $G$ has a positive gap $\Delta>0$ over that of real generation process $H$
:

\begin{align*}
    \mathbb{E}_{Y \sim G(\cdot|X)}[
    \log p(Y|X)
    ]
    -
    \mathbb{E}_{Y \sim H(\cdot|X)}[
    \log p(Y|X)
    ]
    &
    > \Delta.
\end{align*}

This hypothesis states that, conditioned on the input parts of image, the log-likelihood value of the machine-generated remaining parts of image is significantly higher than the human-generated remaining pixels. 
An implication is that
\begin{align*}
    \Delta 
    & \le 
    \mathbb{E}_{Y \sim G(\cdot|X)}[
    \log p(Y|X)
    ]
    -
    \mathbb{E}_{Y \sim H(\cdot|X)}[
    \log p(Y|X)
    ]
    \\
    & \le 
    \| \log p(\cdot|X) \|_{\infty}
    \cdot
    d_{\mathrm{TV}}(G,H)
    \\
    & \le 
    \| \log p(\cdot|X) \|_{\infty}
    \cdot 
    \sqrt{\frac{1}{2}d_{\mathrm{KL}}(G,H)}.
\end{align*}
The second inequality holds due to the definition of the total-variation distance; the third inequality holds due to Pinsker's inequality. When there is no ambiguity, we omit the parenthesis and condition, denote $G(\cdot|X)$ as $G$ and the same for $H$.

\section{Experiments}

\subsection{Experiments Setup}

\noindent\textbf{Datasets.} We consider a variety of generative models, including 
Guided diffusion~\cite{dhariwal2021diffu},
the Latent Diffusion Model (LDM)\cite{rombach2022high}, Glide\cite{nichol2021glide}, DALL-E~\cite{ramesh2021zero}, and DALL-E 3. For these methods, we use the LAION~\cite{schuhmann2021laion} dataset as the real class, while fake images are generated based on the corresponding text descriptions from LAION.

Following the data setup in~\cite{ojha2023towards}, LDMs can generate images in various ways. The standard practice involves using a text prompt as input and performing 200 steps of noise refinement (LDM 200). Additionally, images can be generated with guidance (LDM 200 w/CFG) or using fewer steps for faster sampling (LDM 100).

Similarly, we test on different variants of a pre-trained Glide model, which consists of two stages of noise refinement. The standard approach uses 100 steps to create a low-resolution image at 64 x 64 pixels, followed by 27 steps to upscale the image to 256 x 256 pixels (Glide 100-27). We also consider two other variants: Glide 50-27 and Glide 100-10, which differ in the number of refinement steps used in the two stages.

\begin{table*}[t!]
\centering
  {\small
    \centering
    \tabcolsep=0.1cm
    \resizebox{0.9\linewidth}{!}{
    \begin{tabular}{cc  c ccc ccc c c}
    \toprule
        \multirow{2}{*}{\shortstack[c]{\textbf{Detection method}}} & \multirow{2}{*}{\textbf{Variant}}  &\multirow{2}{*}{\textbf{Guided}} & \multicolumn{3}{c}{\textbf{LDM}} & \multicolumn{3}{c}{\textbf{Glide}} & \multirow{2}{*}{\textbf{DALL-E}} & \textbf{Average} \\
    &  & & \shortstack[c]{\textbf{200}\\\textbf{steps}} & \shortstack[c]{\textbf{200}\\\textbf{w/ CFG}} & \shortstack[c]{\textbf{100}\\ \textbf{steps}} & \shortstack[c]{\textbf{100}\\ \textbf{27}} & \shortstack[c]{ \textbf{50}\\ \textbf{27}} & \shortstack[c]{\textbf{100}\\ \textbf{10}} & & \shortstack[c]{\textbf{mAP}} 
    \\ 
    \midrule
\multirow{3}{*}{\shortstack[c]{\textbf{Trained}\\\textbf{deep network} \cite{wang2020cnn}}} & Blur+JPEG (0.1) & 73.72 & 70.62 & 71.0 & 70.54 & 80.65 & 84.91 & 82.07 & 70.59 & 75.51 \\
& Blur+JPEG (0.5)  & 68.57 & 66.0 & 66.68 & 65.39 & 73.29 & 78.02 & 76.23 & 65.93 & 70.01 \\
& ViT:CLIP (B+J 0.5)  & 55.74 & 52.52 & 54.51 & 52.2 & 56.64 & 61.13 & 56.64 & 62.74 & 56.52 \\ 
\midrule
\multirow{2}{*}{\shortstack[c]{\textbf{Patch}\\\textbf{classifier} \cite{chai2020makes}}} & ResNet50-Layer1   & 70.05 & 87.84 & 84.94 & 88.1 & 74.54 & 76.28 & 75.84 & \textbf{77.07} & 79.33 \\
& Xception-Block2 & 75.03 & 87.1 & \textbf{86.72} & 86.4 & 85.37 & 83.73 & 78.38 & 75.67 & 82.30\\
\midrule
\shortstack[c]{\textbf{Freq-spec} \cite{zhang2019detecting}} & CycleGAN & 57.72 & 77.72 & 77.25 & 76.47 & 68.58 & 64.58 & 61.92 & 67.77 & 69.00 \\
\midrule
\shortstack[c]{\textbf{Ours}} & Stable Diffusion & \textbf{92.97} & \textbf{89.40} & 82.84 & \textbf{90.41} & \textbf{87.75} & \textbf{86.78} & \textbf{86.75} & 75.98 & \textbf{86.61} \\
    \bottomrule
    \end{tabular}}
    }
    \vspace{-2.5mm}
    \caption{Fake image detection results, evaluated with the Average Precision (AP) metric. Results are presented for different generative models (Guided Diffusion, LDM, Glide, and DALL-E) under varying configurations, such as sampling steps and guidance levels. 
    The ``Average mAP'' column represents the mean AP across all generative model variants. 
    }
    \label{tab:ap}
\end{table*}

\noindent\textbf{Baselines.} We consider several strong baseline methods:
\emph{i}) Trained Deep Network \cite{wang2020cnn}: This method uses a ResNet-50 \cite{he2016deep} pre-trained on ImageNet, fine-tuned on ProGAN's real and fake images to make real/fake decisions using binary cross-entropy loss;
\emph{ii}) Patch Classifier~\cite{chai2020makes}: This approach trains a similar classification network, but operates at the patch level instead;
\emph{iii}) Freq-Spec~\cite{zhang2019detecting}: This technique trains a classification network on the frequency spectrum of real and fake images.

\noindent\textbf{Metrics.} We use the metrics of Average Precision (AP), Area Under The ROC Curve (AUROC) score, and FPR95 to evaluate the detection quality.  The threshold for the detector is selected based on the fake data when 95\% of fake test data points are declared as fake.

\noindent\textbf{Experimental details.} In detecting AI-generated content, we consider two realistic scenarios: the white-box scenario, where we have access to the target generative model, and the black-box scenario, where we do not.
In the white-box scenario, we directly perform recovery-based fake detection using the generative model that produced the fake images. In the black-box scenario, we employ a strong stable diffusion model as a surrogate in our experiments. Additional experimental details can be found in the Appendix~\ref{app:exp-details}.

\subsection{Main Results and Analysis}\label{exp:main}

We begin by comparing our approach against baseline methods for identifying fake images generated by various models. In addition, we perform ablation studies to analyze the impact of different components of our approach.

\noindent\textbf{Compared against baseline methods for fake image detection.} Table~\ref{tab:ap} presents the average precision (AP) of various methods for detecting AI-generated content across different generative models. While the trained classifier baseline achieves high accuracy for GAN variants~\cite{wang2020cnn}, its performance significantly drops for modern diffusion-based generative models. This trend remains consistent even when switching the backbone from standard deep neural networks to the CLIP:ViT model, which performs slightly worse. These results suggest that detection accuracy may suffer from overfitting when using models with larger capacities. Performing classification at the patch level~\cite{chai2020makes} or utilizing the frequency domain~\cite{zhang2019detecting}, does not achieve consistent detection performance. This indicates that learning patterns from small image patches alone is insufficient to address the problem. Current fake detection baselines struggle to reliably identify complex generative content.

On the other hand, our approach demonstrates significantly higher average precision in identifying fake images. By using Stable Diffusion as a surrogate model, our method outperforms baseline approaches—such as trained deep networks~\cite{wang2020cnn}, patch classifiers~\cite{chai2020makes}, and frequency-based methods~\cite{zhang2019detecting}—across a diverse range of advanced diffusion-based generative model datasets. Our method maintains a high mean Average Precision (mAP) of approximately 86.81\% for fake content detection, clearly surpassing the performance of these baselines. This result highlights the effectiveness of our approach in recovering from masked images, particularly for the challenging task of black-box AI-generated content detection.

\begin{figure*}[!t]
\vspace{-0.2cm}
\centering
\begin{subfigure}[b]{0.24\textwidth}
    \includegraphics[width=\textwidth]{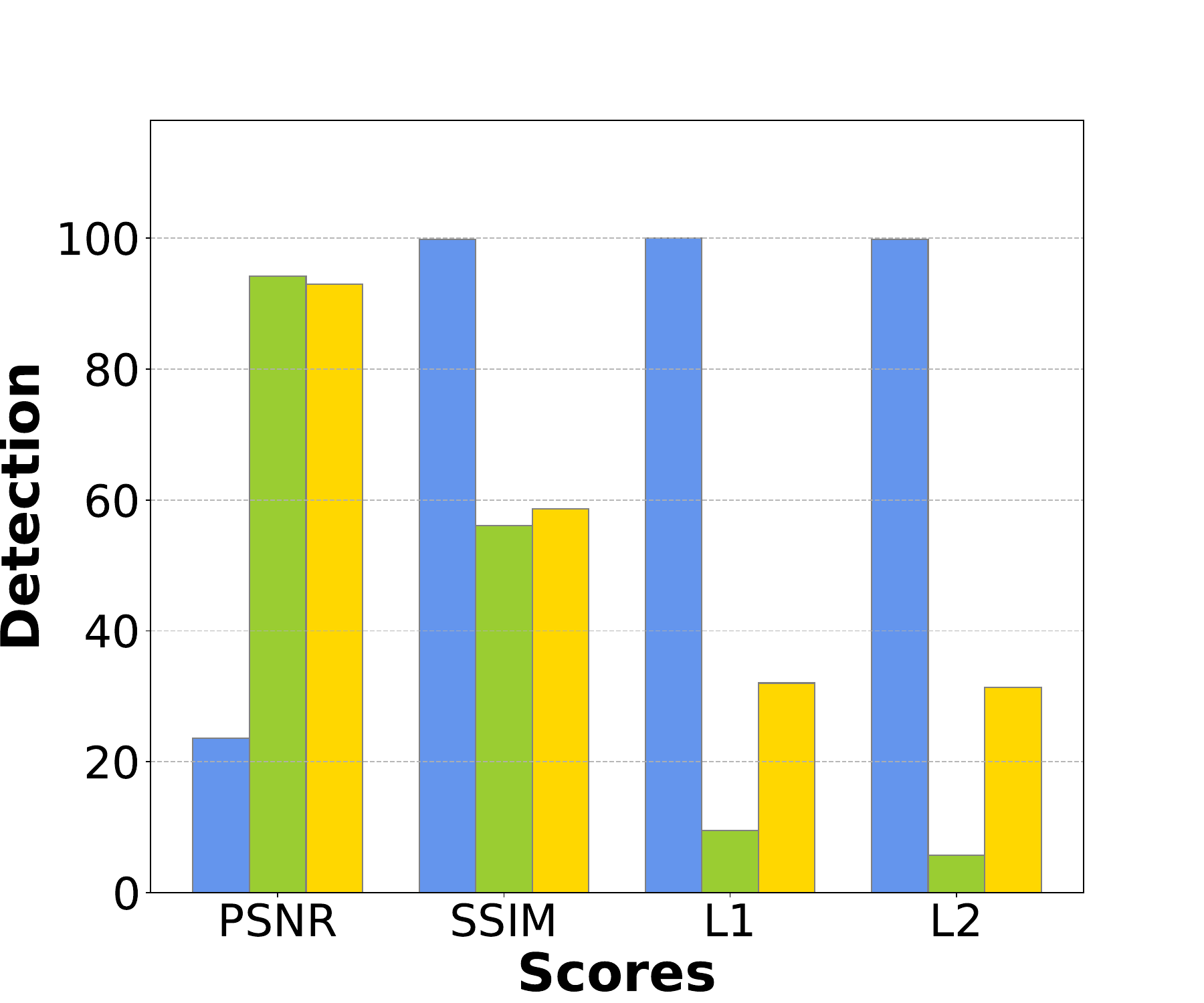}
    \caption{Guided}
    \label{guided-result}
\end{subfigure}%
\begin{subfigure}[b]{0.24\textwidth}
    \includegraphics[width=\textwidth]{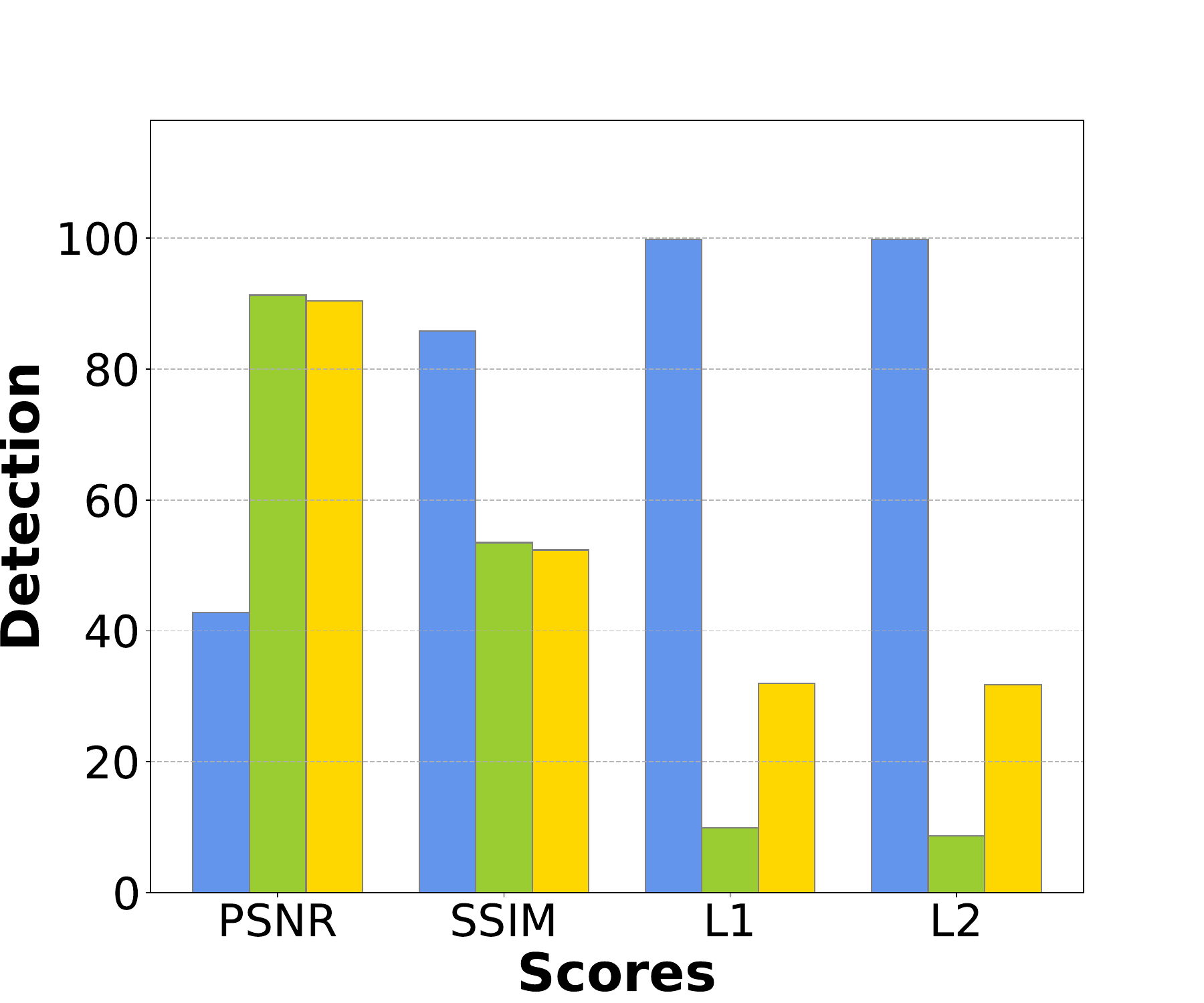}
    \caption{LDM}
    \label{ldm100-result}
\end{subfigure}%
\begin{subfigure}[b]{0.24\textwidth}
    \includegraphics[width=\textwidth]{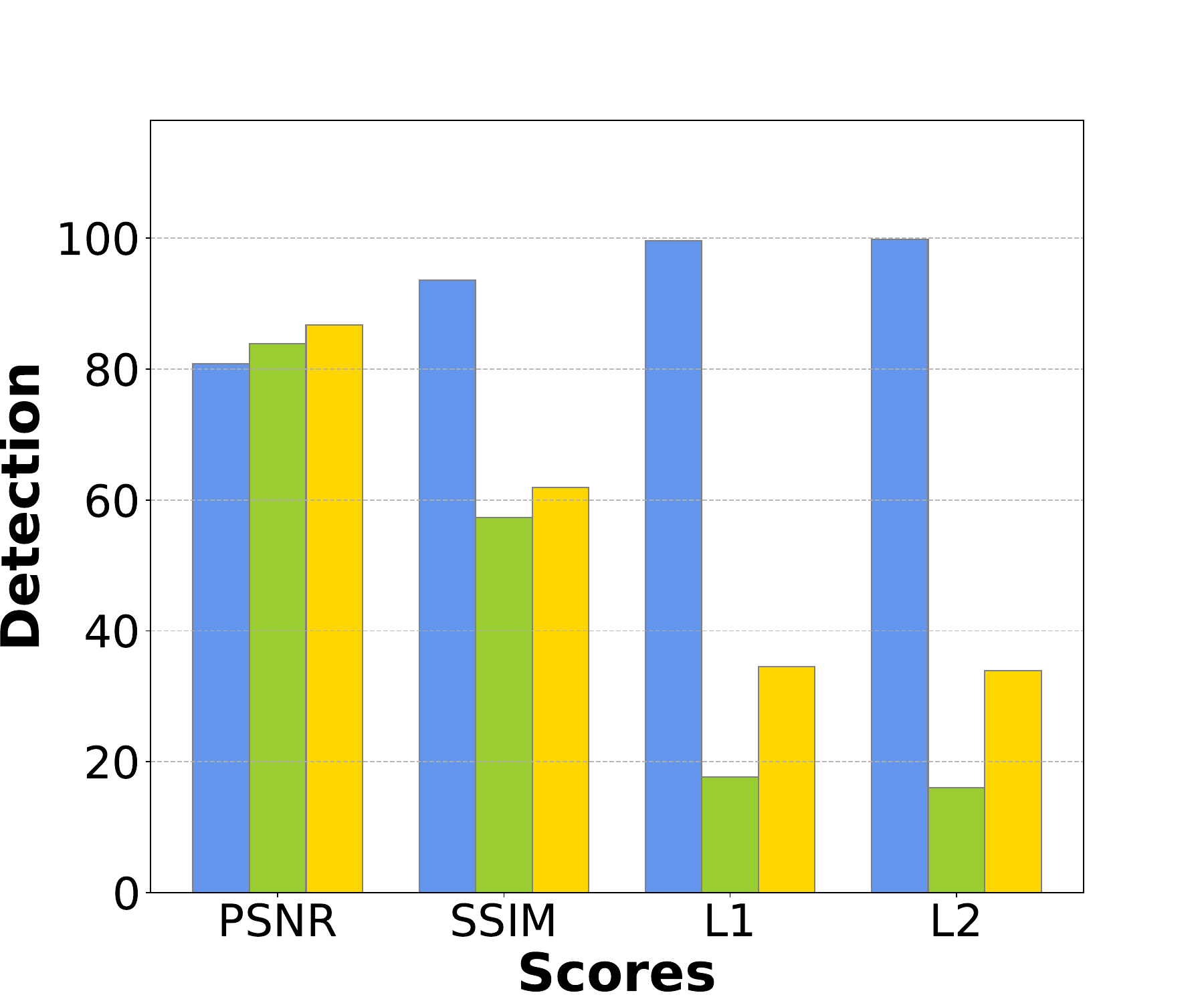}
    \caption{Glide}
    \label{glide-result}
\end{subfigure}%
\begin{subfigure}[b]{0.24\textwidth}
    \includegraphics[width=\textwidth]{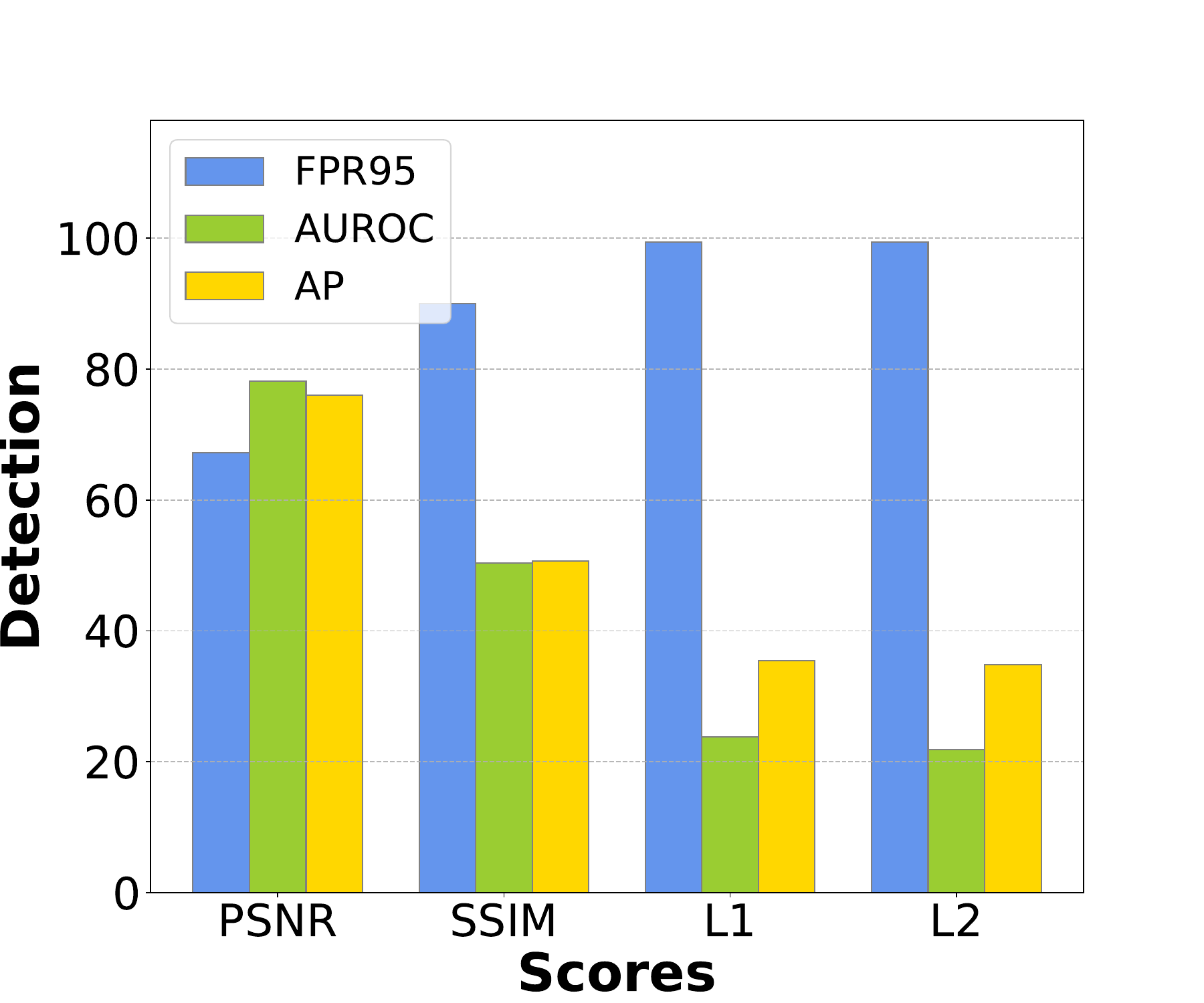}
    \caption{DALL-E}
    \label{dalle-result}
\end{subfigure}%
\caption{\small 
Evaluation on differen scores and evaluation metrics. The PSNR scores here show better detection accuracy in terms of FPR95, AUROC, and AUPR. This trend is consistent across different datasets, including (a) Guided, (b) LDM, (c) Glide, (d) DALL-E. 
} 
\label{fig:dataset}\vspace{-0.4cm}
\end{figure*}

\noindent\textbf{New dataset for advanced generative models DALL-E 3.} 
Recent commercial tools (e.g., DALL-E 3) have made remarkable strides in synthesizing photorealistic images. However, DALL-E 3 is currently only accessible via APIs, and there are very few benchmarks focused on detecting fake content from this new model. Additionally, most existing AI-generated content detection benchmarks lack paired images representing real and fake distributions. Typically, these benchmarks consist of randomly selected images from different generative models, which may introduce confounding factors, leading to shortcut learning in fake detection and hindering their generalization capabilities.

In this work, we design a new benchmark specifically for DALL-E 3-based content detection. We begin by collecting a set of high-resolution, high-quality real images covering a wide range of real-world categories, referencing the LAION dataset. Next, we generate detailed captions for these real images using the LLaVA model. Based on these captions, we then generate corresponding images using DALL-E 3. This process allows us to create paired data for evaluating fake image detection, effectively eliminating confounding factors. This paired dataset provides a more challenging and robust evaluation setting, offering a better measure of detection methods' performance.

Our proposed dataset is highly challenging, with existing methods showing a significant drop in performance compared to traditional benchmarks. In summary, the newly proposed DALL-E 3 fake detection dataset is more challenging due to the model's photorealistic generation capabilities and the inclusion of paired real-fake images, which reduces unrelated factors in detection tasks. This benchmark provides a valuable resource for advancing future research in AI content detection.

\begin{table}
\centering
\begin{tabular}{cc|ccc}
\toprule
\textbf{Method} & \textbf{Scores} 
& \textbf{FPR}$\downarrow$ & \textbf{AUROC}$\uparrow$ &\textbf{AP}$\uparrow$   \\
\midrule
\multirow{2}*{\textbf{\shortstack[c]{Ours w/o\\fine-tuning}}}& PSNR & 47.90 & 87.84 & 86.74   \\
& SSIM & 100 & 45.28 &  44.36   \\
\midrule
\multirow{2}*{\textbf{\shortstack[c]{Ours with\\fine-tuning}}} & PSNR  & 23.60 & 94.19 & 92.97  \\
  & SSIM & 99.80 & 56.13 &  58.60  \\
\bottomrule
\end{tabular}%
\caption{Experiments study on comparing the performance of the model after applying fine-tuning versus without parameter-efficient fine-tuning. The study utilized datasets tailored for guided diffusion models, with Stable Diffusion serving as the base model and testing on guided diffusion datasets.}
\label{tab:ablation-stable}
\end{table}

\subsection{Ablation Studies}\label{exp:ablation}

\noindent\textbf{Ablations on different components.}
As shown in Table~\ref{tab:ablation-stable}, we experiment with a variant of our recovery-based black-box detection method. We observe that the inpainting-based recovery approach, without parameter-efficient fine-tuning, results in a 47.90\% FPR and an 86.74\% average precision on the guided diffusion datasets. However, our approach, which leverages a surrogate model (e.g., stable diffusion) with parameter-efficient fine-tuning on a small set of examples from the target generative model (e.g., guided diffusion), significantly enhances fake detection accuracy for the unknown target model. This method achieves a 23.60\% FPR and a 92.97\% average precision, representing a 24.3\% reduction in FPR. These results highlight the importance of parameter-efficient fine-tuning for effective black-box detection when only API access to the target model is available, without knowledge of its internal weights.

We further conduct an ablation study using guided diffusion as surrogate model, testing it on various diffusion datasets, including guided diffusion, DALL-E, and GLIDE. With appropriate scoring measures, we observe that when the guided diffusion model is tested on guided datasets, it demonstrates significantly better detection performance, achieving an FPR of 10.80\%, an AUROC of 97.18\%, and an average precision of 96.69\%. In contrast, the detection performance on DALL-E yields an FPR of 67.80\% and an AUROC of 76.75\%, while on GLIDE, it shows an FPR of 77.70\% and an AUROC of 85.67\%. These results are notably lower compared to detecting guided diffusion images.

This indicates that diffusion models are better at distinguishing images that are closer to the distribution they are trained on, making recovery easier compared to real images. As a result, the discrepancy between fake and real images is more pronounced, leading to improved detection accuracy. This also supports the effectiveness of our recovery-based fake detection method in a white-box scenario. Specifically, if we have access to the target diffusion model, we can use recovery scores for fake detection without the need for fine-tuning, demonstrating the flexibility of our method for both white-box and black-box detection scenarios.

\begin{table}
\centering
\begin{tabular}{cc|ccc}
\toprule
\textbf{Method} & \textbf{Scores} 
& \textbf{FPR}$\downarrow$ & \textbf{AUROC}$\uparrow$ &\textbf{AP}$\uparrow$   \\
\midrule
\multirow{2}*{\textbf{Guided}}& PSNR & 10.80 &97.18  & 96.69   \\
& SSIM & 76.70 & 67.25 & 63.81    \\
\midrule
\multirow{2}*{\textbf{DALL-E}} & PSNR  & 67.80 & 76.52 & 73.18  \\
  & SSIM & 93.40 & 60.01 & 60.95   \\
\midrule
\multirow{2}*{\textbf{Glide}} & PSNR & 77.70 & 85.67 & 87.75 \\
 & SSIM & 95.50 & 48.34 & 48.25  \\
\bottomrule
\end{tabular}
\caption{Evaluation was conducted on different datasets using guided diffusion as the base model. The experiments were performed without parameter-efficient fine-tuning.}
\label{tab:ablation-guided}
\end{table}

\noindent\textbf{Ablations on different scores.}
Figure~\ref{fig:dataset} and Figure~\ref{fig:masks} present ablations using different metrics—PSNR, SSIM, L1, and L2—for measuring the discrepancy between real and fake images across various datasets. We observe that PSNR demonstrates significantly better performance in detecting fake images compared to the other three metrics. For instance, when using stable diffusion as the surrogate model and guided diffusion as the target model, the AUROC achieved with the PSNR is 94.19\%, whereas SSIM yields only 56.13\% AUROC under the same setting—a direct improvement of 38.06\%. This trend consistently holds across different surrogate models, target models, and masks.

These highlight that selecting an appropriate metric, such as PSNR, is crucial for achieving high detection performance, while the use of less suitable metrics can severely hinder the model's ability to identify fake images. This observation also suggests that designing more tailored metrics could further enhance recovery-based detection methods, offering promising directions for future research.

\noindent\textbf{Ablations on different masks.} We evaluate different variants of mask types, as shown in Figure~\ref{fig:masks}, with visualizations provided in Appendix~\ref{app:masks}. The choice of mask type influences the performance of recovery-based black-box detection. Notably, the genhalf mask demonstrates slightly better fake detection accuracy across all four metrics compared to thick-type masks. This improvement may be attributed to the larger masked region in the genhalf mask, which increases the recovery area used to compute the discrepancy.

\begin{figure}[t]
\vspace{-0.2cm}
\centering
\begin{subfigure}[b]{0.21\textwidth}
    \includegraphics[width=\textwidth]{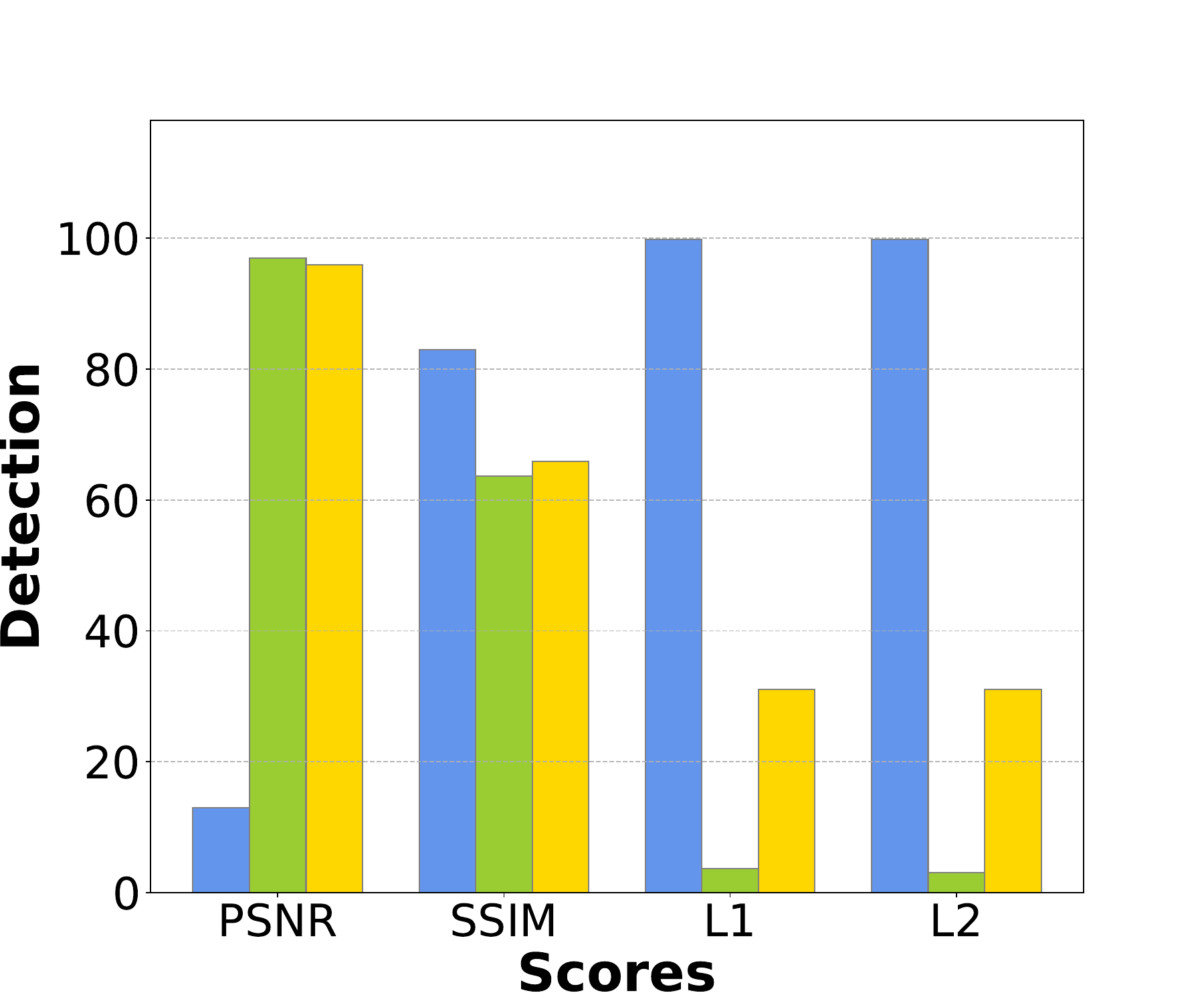}
    \caption{Thick Mask}
    \label{thick-mask}
\end{subfigure}%
\begin{subfigure}[b]{0.21\textwidth}
    \includegraphics[width=\textwidth]{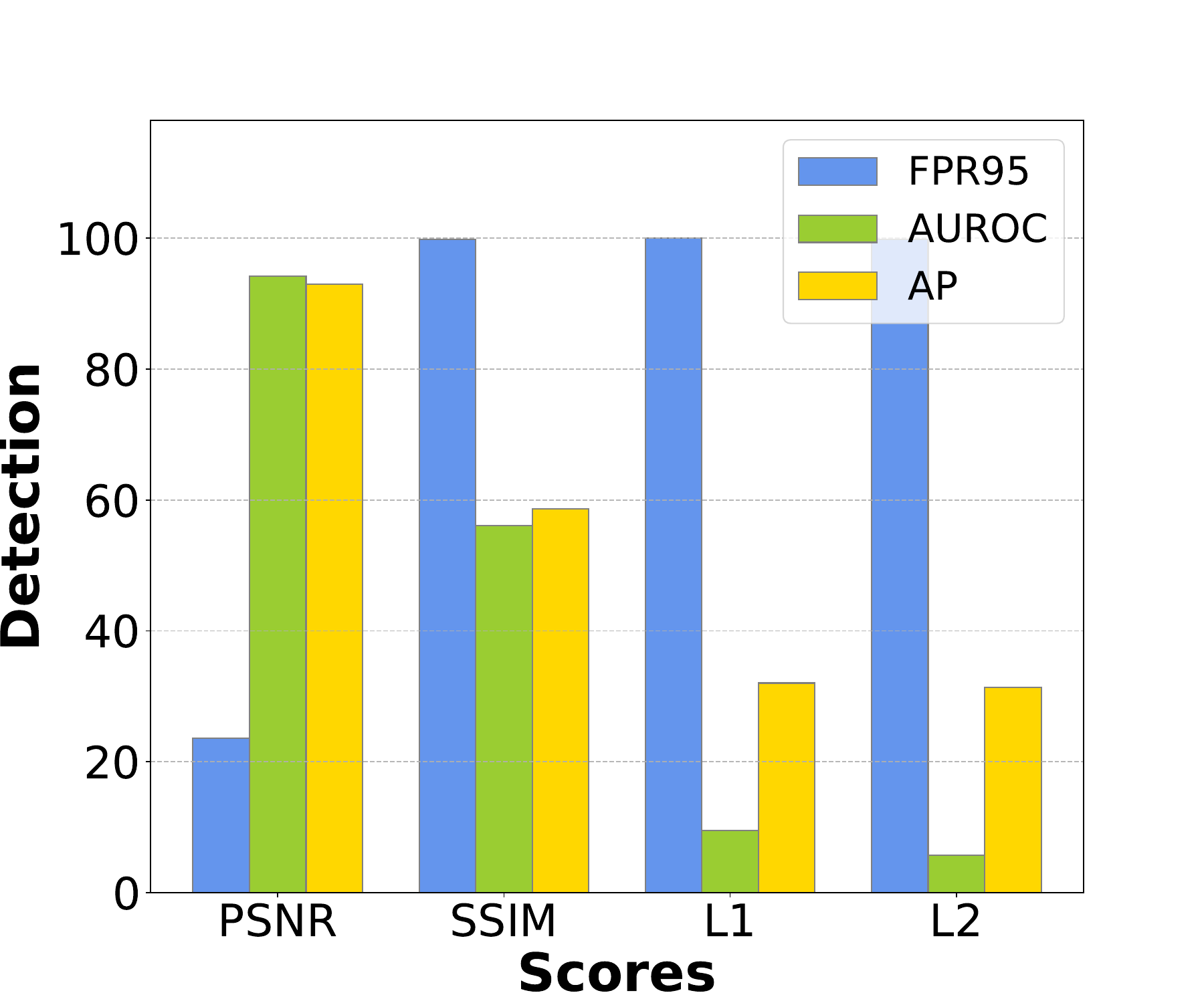}
    \caption{Genhalf Mask}
    \label{genhalf-mask}
\end{subfigure}%
\caption{\small 
Different evaluation scores demonstrate significant differences in fake detection performance. The PSNR scores here show better detection accuracy in terms of FPR95, AUROC, and AUPR. This trend is consistent across different types of masks used for recovery, such as (a) thick masks and (b) genhalf masks. 
}
\label{fig:kde}
\end{figure}

\noindent\textbf{Visualization and qualitative analysis.} We visualize the score distributions in Figure~\ref{fig:kde} (a) and (b) for the Guided vs. Laion and DALL-E vs. Laion settings. There are two key obervations: first, the PSNR scores for fake data are consistently higher than those for real data (Laion), indicating that fake images are better recovered with higher quality using our recovery-based black-box fake detection model. Additionally, the score distributions for Guided vs. Laion show better separation compared to DALL-E vs. Laion. This suggests that our fake detection method is more effective when the generative model's distribution aligns closely with the target test model. This finding highlights the importance of parameter-efficient fine-tuning steps for detecting fake images from closed-source, advanced generative models.

\subsection{Human preference evaluation.}\label{exp:human}
We conducted a human preference evaluation for AI-generated content detection, focusing on comparing images generated by advanced AI models, such as DALL-E 3, to real or human-created images. Respondents were asked to carefully observe the provided images and identify which ones they believed were generated by DALL-E 3. The evaluation consisted of 100 questions, with images randomly selected from our curated DALL-E 3 fake detection dataset.

Our observations are as follows: \emph{i}) The average accuracy of human respondents was 72.33\%, indicating that DALL-E 3-generated images are challenging to distinguish even for human eyes. \emph{ii}) Distinguishing between DALL-E 3-generated and human-created content was notably harder for categories like art paintings and cartoons compared to realistic photographs such as landscapes and human figures. This could be because people are more familiar with real-world scenes, making it easier to identify subtle inconsistencies in those contexts. \emph{iii}) The accuracy of individual respondents ranged from 59\% to 89\%, highlighting that people's ability to differentiate between AI-generated and real images varies based on their backgrounds, areas of expertise, and familiarity with modern commercial AI tools.

\begin{figure}[t]
\centering
\begin{subfigure}[b]{0.21\textwidth}
    \includegraphics[width=\textwidth]{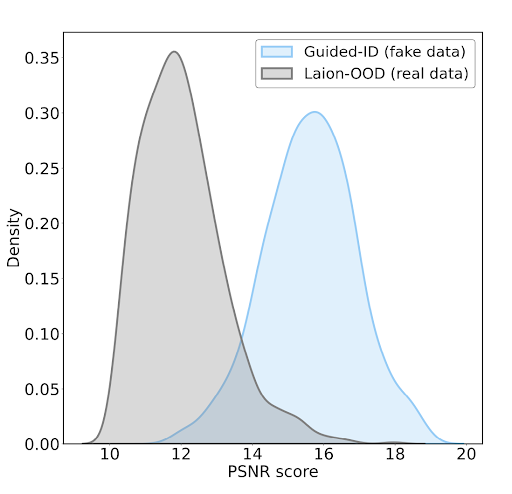}
    \caption{Guided vs Laion}
    \label{guided}
\end{subfigure}%
\hspace{1mm}
\begin{subfigure}[b]{0.21\textwidth}
    \includegraphics[width=\textwidth]{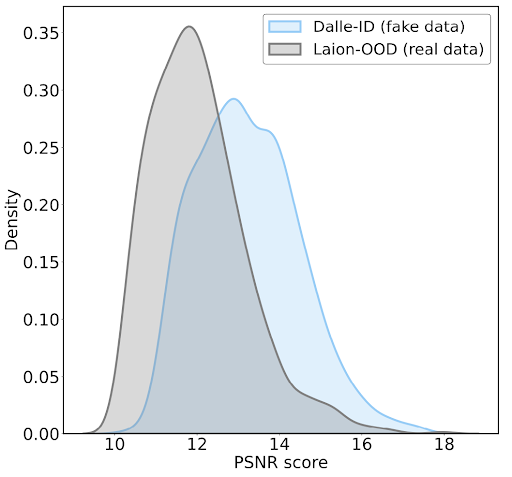}
    \caption{DALL-E vs Laion}
    \label{dalle}
\end{subfigure}%
\caption{\small 
KDE visualization without fine-tuning steps. The surrogate model is a guided diffusion model. The images generated by the guided diffusion model are better identified, as they are more aligned with the distribution of the base model.
}
\label{fig:kde}\vspace{-0.4cm}
\end{figure}

\section{Conclusion}
In this work, we introduce a novel recovery-based black-box detection framework for distinguishing AI-generated content from real images. Our method leverages the differences in the recovery quality between real and synthetic images from the masked regions. By aligning the distribution of a surrogate model with that of a black-box target model through parameter-efficient fine-tuning, we achieved significant improvements in detection performance.
Extensive experiments across various diffusion models demonstrate the effectiveness of our method. Notably, we observe a specific scoring measure proved superior in detecting fake images compared to other metrics. This highlights the critical role of selecting appropriate discrepancy measures in enhancing detection accuracy.
Our findings underscore the need for robust, cost-efficient detection methods, particularly in scenarios where access to model internals is restricted. The success of our recovery-based strategy also opens up new avenues for developing more tailored scoring metrics and recovery techniques to address evolving capabilities of advanced generative models.

\clearpage

{
    \small
    \bibliographystyle{ieeenat_fullname}
    \bibliography{main}
}

\clearpage

\appendix

\section{Theoretical Analysis}\label{app:theory}

\subsection{Is it always possible to distinguish between Generated content and real images?}
The recent work explores the detection of AI-generated content by analyzing the AUROC for any detector \( D \). It leverages Le Cam's lemma \citep{le2012asymptotic,wasserman2023lecture}, which states that for any distributions \( G \) and \( H \), given an observation \( s \), the minimum sum of Type-I and Type-II error probabilities in testing whether \( s \sim G \) or \( s \sim H \) is equal to \( 1 - d_{\text{TV}}(G, H) \), where \( d_{\text{TV}} \) denotes the total variation distance between the two distributions. 
This result can be interpreted as:
\begin{align}\label{ROC_upper}
    \text{TPR}_{\gamma} \leq \min\{\text{FPR}_{\gamma} + d_{\text{TV}}(G, H),1\},
\end{align}

where $\text{TPR}_{\gamma}\in[0,1]$. The upper bound in \eqref{ROC_upper} is leveraged in one of the recent work \citep{sadasivan2023can} to derive \texttt{AUROC} upper bound $\texttt{AUC} \leq \frac{1}{2} + d_{\text{TV}} (G, H) - \frac{d_{\text{TV}} (G, H)^2}{2}$ which holds for any $D$. 
This upper bound led to the claim of impossibility results for reliable detection of AI-Generated content when $d_{\text{TV}} (G, H)$ is approaching 0. The upper bound in \eqref{ROC_upper} is also interpreted as either certain real images will be detected falsely as AI-generated content will not be detected reliably when $d_{\text{TV}} (G, H)$ is small. 
However, as discussed in Sec. \ref{sec:method}, the {Likelihood-Gap Hypothesis} guarantees that the difference between the two distributions is significant enough ($d_{\mathrm{TV}}(G,H)$ or $d_{\mathrm{KL}}(G,H)$ is greater than some positive gap). This implies it is always possible to distinguish between real and machines.

\subsection{Principled Choice of $K$}\label{app:theory_k}
\label{sec:kselect}
In Sec.~\ref{sec:method}, we propose the \textbf{Likelihood-Gap Hypothesis}, which posits that the expected log-likelihood of the machine generation process \( G \) exceeds that of the human generation process \( H \) by a positive gap, \( \Delta > 0 \). 
To exploit this difference between the distributions, we introduce a distance function \( D(Y, Y') \) that quantifies the similarity between two images \( Y \) and \( Y' \). This distance function can also be interpreted as a kernel function used in kernel density estimation.

By re-prompting the masked pixels, we can evaluate how closely the remaining pixels \( Y_0 \) align with the machine-generated distribution:
   $ \hat{D}(Y_0, \{ Y_k \}_{k \in [K]})
     :=
    \frac{1}{K}
    \sum_{k=1}^{K}
    D(Y_0, Y_k)$,
where $K$ is the number of times of re-prompting. 

Similar to the kernel density estimation, we can use this quantity and some threshold to determine whether to accept or reject that $S \sim G$. Under certain assumptions, this estimator enjoys $n^{-1/2}$-consistency via Hoeffding's argument. In the following, we provide a formal argument.

\begin{assumption}
Suppose we have a given human-generated content $[X, Y_0] \in \mathrm{supp}(h)$ and a machine-generated remaining pixels $\tilde{Y}_0$, consider the random variable $D(Y_0,Y')$ and where $Y'$ is sampled by re-prompting given $X$, that is $Y' \sim G(\cdot | X)$. We assume $D(Y_0,Y')$ and $D(\tilde{Y}_0,Y')$ are $\sigma$-sub-Gaussian. We also assume that the distance gap is significant:
    $\mathcal{E}_{Y' \sim G}[ D(Y_0, Y') | X]
    -
    \mathcal{E}_{Y' \sim G}[ D(\tilde{Y}_0, Y') | X]
    > \Delta$.
\end{assumption}

From this assumption, we can derive that it suffices to re-prompt $\Omega \big( 
\frac{\sigma \log(1/\delta)}{\Delta^2} \big)$ times.
\begin{proof}
Note that $\mathcal{E}[\hat{D}] = \mathcal{E}[D]$ and the distribution is sub-Gaussian. By Hoeffding's inequality, we have that with probability at least $1 - \delta$,
\begin{align*}
    \bigg| 
    \frac{1}{K}
    \sum_{k=1}^{K}
    D(Y_0, Y_k)
    - 
    \mathcal{E}_{Y' \sim G}[ D(Y_0, Y') | X]
    \bigg|
    & \le 
    \sqrt{\frac{\sigma \log(\delta / 2)}{K}}.
\end{align*}
Similarly, we have that with probability at least $1 - \delta$,
\begin{align*}
    \bigg| 
    \frac{1}{K}
    \sum_{k=1}^{K}
    D(\tilde{Y}_0, Y_k)
    - 
    \mathcal{E}_{Y' \sim G}[ D(\tilde{Y}_0, Y') | X]
    \bigg|
    & \le 
    \sqrt{\frac{\sigma \log(\delta / 2)}{K}}.
\end{align*}
By the union bound, we have that with probability $1-2\delta$,
\begin{align*}
    & \frac{1}{K}
    \sum_{k=1}^{K}
    D(Y_0, Y_k)
    -
    \frac{1}{K}
    \sum_{k=1}^{K}
    D(Y_0, Y_k)
    \\
    & 
    >
    \frac{1}{K}
    \sum_{k=1}^{K}
    D(Y_0, Y_k)
    - 
    \mathcal{E}_{Y' \sim G}[ D(\tilde{Y}_0, Y') | X]
    \\&\quad\quad-
    \frac{1}{K}
    \sum_{k=1}^{K}
    D(\tilde{Y}_0, Y_k)
    + 
    \mathcal{E}_{Y' \sim G}[ D(\tilde{Y}_0, Y') | X]
    + \Delta 
    \\
    & \ge 
    \Delta - 2 \sqrt{\frac{\sigma \log(\delta / 2)}{K}}.
\end{align*}\small
If we set $K = \Omega \big( 
\frac{\sigma \log(1/\delta)}{\Delta^2} \big) $, then there is a gap between the real distance and the machine's distance.
\end{proof}

\begin{figure*}
\vspace{-0.4cm}
  \centering
  \includegraphics[width=0.78\textwidth]{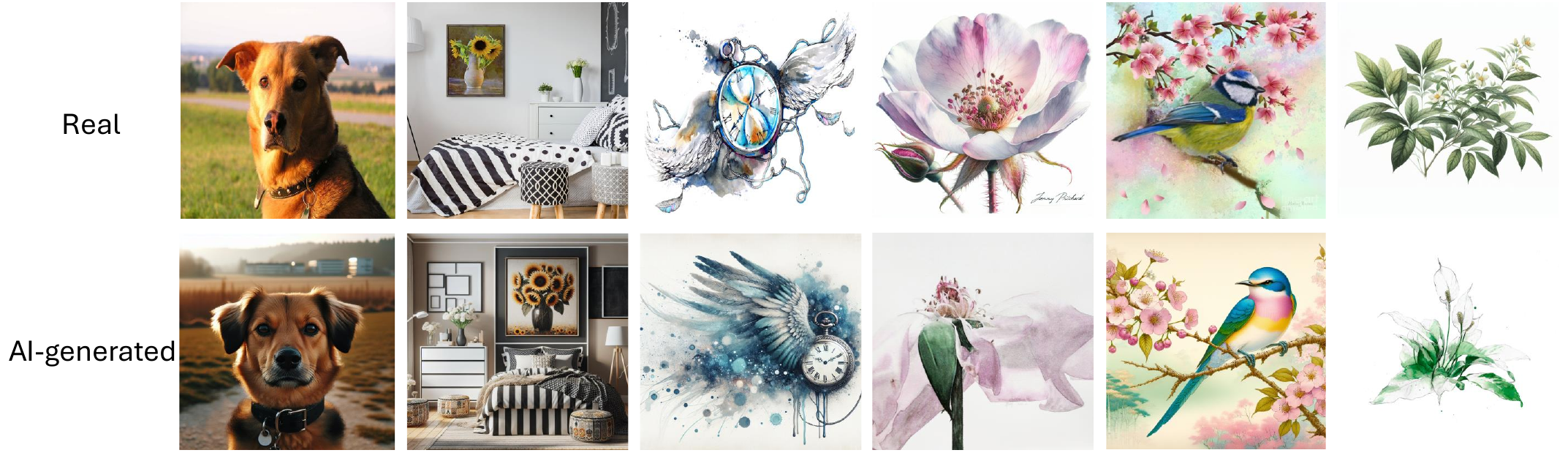}
  \caption{Examples of hard cases for distinction in human evaluations.}\label{fig:hard}
  \vspace{-0.3cm}
\end{figure*}

\section{More Details of Scoring Function $\delta$}\label{app:scores}

In this section, we provide additional details about scoring function $\delta$ including PSNR, SSIM, L1 distance and L2 distance.
Let \( \mathbf{I} \in \mathbb{R}^{h \times w \times c} \) be the original image, where \( h \) and \( w \) are the height and width, respectively, and \( c \) is the number of channels.
Let \( \mathbf{I}' \in \mathbb{R}^{h \times w \times c} \) be the recovered image.
Let \( \text{MAX} \) denote the maximum possible pixel value (e.g., \( 255 \) for 8-bit images).

\noindent\textbf{Peak Signal-to-Noise Ratio (PSNR)} measures the ratio between the maximum possible value of a pixel and the power of the distortion (i.e., Mean Squared Error) between the original and reconstructed images. The Mean Squared Error (MSE) defined as follows:
\[
\text{MSE}(\mathbf{I}, \mathbf{I}') = \frac{1}{h \times w \times c} \sum_{i=1}^{h} \sum_{j=1}^{w} \sum_{k=1}^{c} \left( \mathbf{I}(i, j, k) - \mathbf{I}'(i, j, k) \right)^2.
\]

\noindent The PSNR formula: 
$\text{PSNR}(\mathbf{I}, \mathbf{I}') = 10 \cdot \log_{10} \left( \frac{\text{MAX}^2}{\text{MSE}(\mathbf{I}, \mathbf{I}')} \right)$,
a higher PSNR value indicates a smaller difference between the images, implying better recovery.

\noindent \textbf{Structural Similarity Index (SSIM)} is designed to measure perceptual differences between two images, taking into account luminance, contrast, and structural information. The formula is:
$\text{SSIM}(\mathbf{I}, \mathbf{I}') = \frac{(2 \mu_{\mathbf{I}} \mu_{\mathbf{I}'} + C_1) (2 \sigma_{\mathbf{I} \mathbf{I}'} + C_2)}{(\mu_{\mathbf{I}}^2 + \mu_{\mathbf{I}'}^2 + C_1) (\sigma_{\mathbf{I}}^2 + \sigma_{\mathbf{I}'}^2 + C_2)}$,
where\( \mu_{\mathbf{I}} \) and \( \mu_{\mathbf{I}'} \) are the means of \( \mathbf{I} \) and \( \mathbf{I}' \).
 \( \sigma_{\mathbf{I}}^2 \) and \( \sigma_{\mathbf{I}'}^2 \) are the variances of \( \mathbf{I} \) and \( \mathbf{I}' \).
 \( \sigma_{\mathbf{I} \mathbf{I}'} \) is the covariance between \( \mathbf{I} \) and \( \mathbf{I}' \).
 \( C_1 \) and \( C_2 \) are small constants to stabilize the division.
 The SSIM values range from \(-1\) to \(1\), where \(1\) indicates a perfect match.

\noindent \textbf{L1 distance} measures the absolute difference between corresponding pixels of original and reconstructed images: $L_1(\mathbf{I}, \mathbf{I}') = \frac{1}{h \times w \times c} \sum_{i=1}^{h} \sum_{j=1}^{w} \sum_{k=1}^{c} \left| \mathbf{I}(i, j, k) - \mathbf{I}'(i, j, k) \right|$, where a lower L1 value indicates a smaller difference between the images.

\noindent \textbf{L2 distance} measures the squared difference between the corresponding pixels of the original and reconstructed images: $L_2(\mathbf{I}, \mathbf{I}') = \frac{1}{h \times w \times c} \sum_{i=1}^{h} \sum_{j=1}^{w} \sum_{k=1}^{c} \left( \mathbf{I}(i, j, k) - \mathbf{I}'(i, j, k) \right)^2,$ where a lower L2 value indicates a smaller difference between the images.
L2 distance is related to PSNR as it forms the basis of its calculation.

\section{Additional Experimental details}\label{app:exp-details}

We provide a detailed description of the datasets and model used in this work:

\noindent \textbf{Stable Diffusion}~\cite{rombach2021highresolution} is a text-to-image model based on diffusion techniques. Originating from latent diffusion, its model and weights have been publicly released. Stable Diffusion was trained on pairs of images and captions from LAION-5B\cite{schuhmann2022laion}, an open large-scale dataset for training image-text models.

\noindent \textbf{Guided Diffusion}~\cite{dhariwal2021diffusion} is a diffusion model that uses gradients from a classifier to guide the denoising process during image synthesis. This approach has proven effective for image generation, surpassing GANs in terms of fidelity while maintaining broad distribution coverage.

\noindent \textbf{GLIDE}~\cite{nichol2021glide} is a text-guided diffusion model designed for photorealistic image generation and editing. It employs classifier-free guidance to enhance image quality while maintaining fidelity to text prompts.

\noindent \textbf{LDM}~\cite{rombach2021highresolution} apply diffusion processes in the latent space of pretrained autoencoders rather than directly in high-dimensional pixel space. This approach significantly reduces computational costs while retaining high-quality image synthesis.

\noindent \textbf{DALL-E}~\cite{ramesh2021zero} is an advanced generative model developed by OpenAI for text-to-image synthesis. It creates highly detailed and imaginative images from natural language descriptions, demonstrating strong performance in generating diverse and realistic visuals while enabling creative applications in content generation and design.

\noindent \textbf{DALL-E 3}~\cite{dalle3} is the latest version of OpenAI's text-to-image generative model, offering significant improvements in fidelity, creativity, and alignment with text prompts. It sets a new standard in text-to-image synthesis.

\noindent \textbf{Hardware and software.} Our framework was implemented using PyTorch 2.3.1. Experiments are performed using the RTX A6000.

\section{Visualization of Different Masks}\label{app:masks}

\begin{figure}
  \centering
  \includegraphics[width=0.33\textwidth]{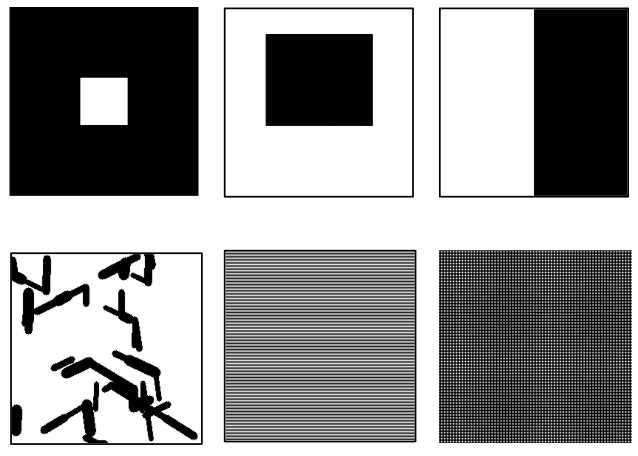}
  \caption{Example of different masks.}\label{fig:masks}
  \vspace{-0.4cm}
\end{figure}

This work supports flexibility with various types of masks. Figure~\ref{fig:masks} illustrates some examples of different masks.

\section{Hard Examples for Distinguishing}\label{App:hard}

Figure~\ref{fig:hard} demonstrates examples where human struggle to accurately differentiate real from AI-generated images.

\end{document}